\documentclass[conference]{IEEEtran}
\usepackage{graphicx}
\usepackage{cite}
\usepackage{amsmath}
\usepackage{pgfplots}
\usepackage{tikz}
\usepackage{subcaption}
\usepackage{caption}
\usepackage{float}

\title{Qwen2.5-32B: Leveraging Self-Consistent Tool-Integrated Reasoning for Bengali Mathematical Olympiad Problem Solving
}
\author{
  \IEEEauthorblockN{Saad Tahmid}
  \IEEEauthorblockA{\textit{Department of Computer Science} \\
                    \textit{Bangladesh University of Engineering and Technology}\\
                    Dhaka, Bangladesh\\
                    Email: saadtahmid1890@gmail.com}
  \and
  \IEEEauthorblockN{Sourav Sarker}
  \IEEEauthorblockA{\textit{Department of Computer Science} \\
                    \textit{Bangladesh University of Engineering and Technology}\\
                    Dhaka, Bangladesh\\
                    Email: souravsarker.2083@gmail.com}

}

\begin{document}

\maketitle

\begin{abstract}
  In this research paper, we present an innovative approach tailored to solving mathematical problems in Bengali, developed for the DL Sprint 3.0 - BUET CSE Fest 2024 Competition. Our methodology harnesses the power of advanced deep learning models, notably the Qwen 2.5 series, with iterative improvements made through prompt engineering, model quantization, and Tool Integrated Reasoning (TIR) to handle complex calculations. Initially, we explored various model architectures, such as fine-tuned Mistral and quantized Qwen models, progressively refining them through translation techniques, RAG (Retrieval-Augmented Generation), and custom dataset curation. Through manual hyperparameter tuning, we optimized parameters like temperature and top-p to improve model adaptability and response accuracy. Additionally, the removal of RAG and careful parameter adjustments further contributed to our final model’s robustness. Our approach demonstrates the potential of advanced NLP techniques in effectively interpreting and solving Bengali mathematical problems.
  
\end{abstract}


\section{Introduction}
The ability to understand and solve mathematical problems is a foundational skill for AI, essential for advancements across fields like science, engineering, and finance. However, while AI models have made strides in various languages, they still face significant challenges when tackling mathematical reasoning in low-resource languages, such as Bengali. This gap becomes particularly evident in tasks involving complex problem-solving and precise calculations. To address this, the DL Sprint 3.0 - BUET CSE Fest 2024 Competition \cite{dlsprint2024} introduced the unique challenge of building an AI model capable of solving mathematical problems in Bengali, targeting issues akin to those in the Bengali Math Olympiad. This competition not only tests participants' technical skills but also aims to push the boundaries of AI's adaptability and performance in Bengali.
Our work contributes to this pioneering effort, focusing on enhancing AI's mathematical reasoning in Bengali through advanced NLP and deep learning techniques. We explore and iteratively refine state-of-the-art models, such as the Qwen series, alongside strategies like prompt engineering, Tool Integrated Reasoning (TIR), and manual hyperparameter tuning to achieve robust problem-solving capabilities. By contributing to this research, we aim to advance AI’s reach into Bengali language processing, ultimately creating models that can assist students, educators, and researchers in tackling complex problems with precision and reliability.

\section{Methodology}

\subsection{Model Selection}
For this task, we aimed to select models capable of efficiently solving mathematical problems in Bengali, considering both performance and computational efficiency. Our initial approach involved using a fine-tuned Mistral 7B model \cite{mCoT2024}, which, despite being a strong general-purpose model, did not deliver the desired accuracy for mathematical reasoning tasks. This led us to explore other models better suited for handling mathematical challenges, particularly in a low-resource language like Bengali.
The Qwen series emerged as the most promising option due to its strong performance on mathematical reasoning benchmarks. The Qwen-32B-Instruct model \cite{qwen2024a}, with impressive scores on the MATH benchmark (83.1) and GSM8K benchmark (95.9) \cite{qwen2024b}, was especially appealing for its capability in solving high-level mathematical problems. Given its robust performance, we chose to focus on the Qwen-2.5 series, including the 7B, 14B, and 32B models, each offering different trade-offs between accuracy and computational demands.
We initially fine-tuned the Qwen-14B-Instruct model for a single epoch, which, though promising, did not provide sufficient improvements in performance. To enhance model efficiency, we implemented VLLM (Variable-Length Language Model) for faster inference, which allowed us to speed up the testing process while maintaining model accuracy. Additionally, we employed model quantization techniques to reduce memory requirements, making the models more practical for large-scale inference tasks.
To further refine performance, we incorporated Tool Integrated Reasoning (TIR), which enabled the model to perform complex calculations using Python. This method improved the model's ability to handle mathematical operations effectively. Moreover, manual hyperparameter tuning of parameters like temperature and top-p helped optimize the model's response accuracy and adaptability.
In summary, after exploring various models, we selected the Qwen series, particularly the Qwen-32B-Instruct model, for its exceptional performance in mathematical reasoning. Combined with techniques like VLLM, TIR, and hyperparameter optimization, we were able to enhance the model’s ability to solve mathematical problems in Bengali effectively.

\subsection{Preprocessing}
In this project, preprocessing was crucial to handling Bengali mathematical problems and enhancing the model's ability to solve them accurately. Given the complexity of understanding Bengali text directly in the initial stage, we leveraged the Qwen-32B-Instruct model to translate Bengali mathematical questions into English for improved processing. This approach enhanced the performance of our model significantly.

\subsection{Prompt Tuning}
Prompt tuning played a key role in optimizing the performance of our model as we experimented with various reasoning techniques, including Chain of Thought (COT), Tool Integrated Reasoning (TIR), RAG, self-consistent TIR, and self-consistent COT.
We initially designed prompts to guide the model through step-by-step reasoning for COT. For TIR, we adapted the prompts to instruct the model to perform calculations using Python tools for more complex problems. With Self-COT and Self-TIR, the prompts were modified to encourage the model to generate multiple reasoning paths and select the most consistent solution. we experimented with Retrieval-Augmented Generation (RAG). The RAG approach was implemented to provide context and improve the quality of answers in both Bengali and translated English questions. However ,  after experimenting with Retrieval-Augmented Generation (RAG), we decided to discontinue its use due to its poor performance, as it did not significantly improve the model’s accuracy compared to other approaches.Additionally, we fine-tuned hyperparameters such as temperature and top\-p to control the diversity and confidence of the model's responses. Lower temperatures were used for more deterministic answers, while higher values promoted creativity.Adjusting top\-p helped the model select the most plausible solutions.These prompt tuning and hyperparameter adjustments, particularly with TIR and Self-TIR, significantly improved the model's ability to solve complex Bengali math problems.

\section{Results}

In this section, we present the performance of our deep learning model using various approaches. The model was evaluated on the public leaderboard of the DL Sprint 3.0, with the baseline score being 28 out of 100. We tested different configurations, including using various versions of the Qwen model, translation, retrieval-augmented generation (RAG), and Tool Integrated Reasoning (TIR).

\begin{table}[H]
  \centering
  \begin{tabular}{|c|c|c|c|c|c|}
       \hline
         \textbf{Translation} & \textbf{RAG} & \textbf{TIR} & \textbf{Self-Consistency} & \textbf{Score} \\
        \hline
         No  & No  & No  & No  & 49 \\
                            Yes(Qwen2.5-14B-Instruct) & No & No  & No & 61 \\
                            No  & Yes & No   & Yes & 48 \\
                            Yes(Qwen2.5-14B-Instruct)  & Yes & No  & Yes & 65 \\
                            Yes(Qwen2.5-32B-Instruct)  & Yes & No  & Yes & 70 \\
                            No  & No & No  & Yes & 66 \\
                            Yes(Qwen2.5-32B-Instruct)  & No & No  & Yes & 73 \\
                                                                 &     & No  & Yes & 77 \\
        \hline
  \end{tabular}
  \caption{ Performance of Qwen2.5-32B-Instruct with Different Configurations}
  \label{table:model_performance}
\end{table}

\begin{table}[H]
  \centering
  \begin{tabular}{|c|c|c|c|c|}
       \hline
         \textbf{Temperature} & \textbf{Top\_p} & \textbf{Number of Candidates} & \textbf{Inference Time(s)} & \textbf{Score} \\
        \hline
         0.2  & 0.9  & 4  & 7844.8  & 68 \\
         0.35  & 0.775  & 4  & 7327.0  & 72 \\
         0.4  & 0.8  & 4  & 7391.5  & 77 \\
         0.4  & 0.8  & 10  & 19753.3  & 77 \\
         0.3  & 0.75  & 4  & 7579.1  & 76 \\
         0.3  & 0.75  & 5  & 9721.5  & 74 \\
         0.3  & 0.75  & 3  & 5436.7 & 71 \\
         0.7  & 0.8  & 4  & 7567.7  & 70 \\
        \hline
  \end{tabular}
  \caption{Hyperparameters and performance of the model Qwen2.5-32B-Instruct(No RAG, Translation,TIR)}
  \label{table:hyperparameter}
\end{table}

\begin{table*}[ht]
  \centering
  \begin{tabular}{|c|c|c|c|c|c|}
       \hline
         \textbf{Model} & \textbf{Translation} & \textbf{TIR} & \textbf{Self-Consistency} & \textbf{Score} \\
        \hline
         Deepseek-math-7b-Instruct(Baseline) & No  & No  & Yes  & 28 \\
                            mCot(Fine-tuned Mistral 7B) & No  & No  & No  & 15 \\
                            Qwen2.5-14B-Instruct & No  & No  & No & 40 \\
                            Qwen2.5-Math7B-Instruct & No  & Yes  & No & 39 \\
                            & Yes  & Yes  & No & 48 \\
                            &       & No  & No & 59 \\
                            NuminaMath-7b-TIR(Finetuned Deepseek-math-7b) & Yes(Qwen2.5-14B-Instruct)  & Yes  & Yes  & 66\\
                            & Yes(Qwen2.5-32B-Instruct)  & Yes  & Yes  & 68\\
                            
        \hline
  \end{tabular}
  \caption{Performance of other tried models with Different Configurations}
  \label{table:model_performance2}
\end{table*}

\begin{figure*}[ht]
\vspace{2em}
  \centering
  \includegraphics[width=\textwidth]{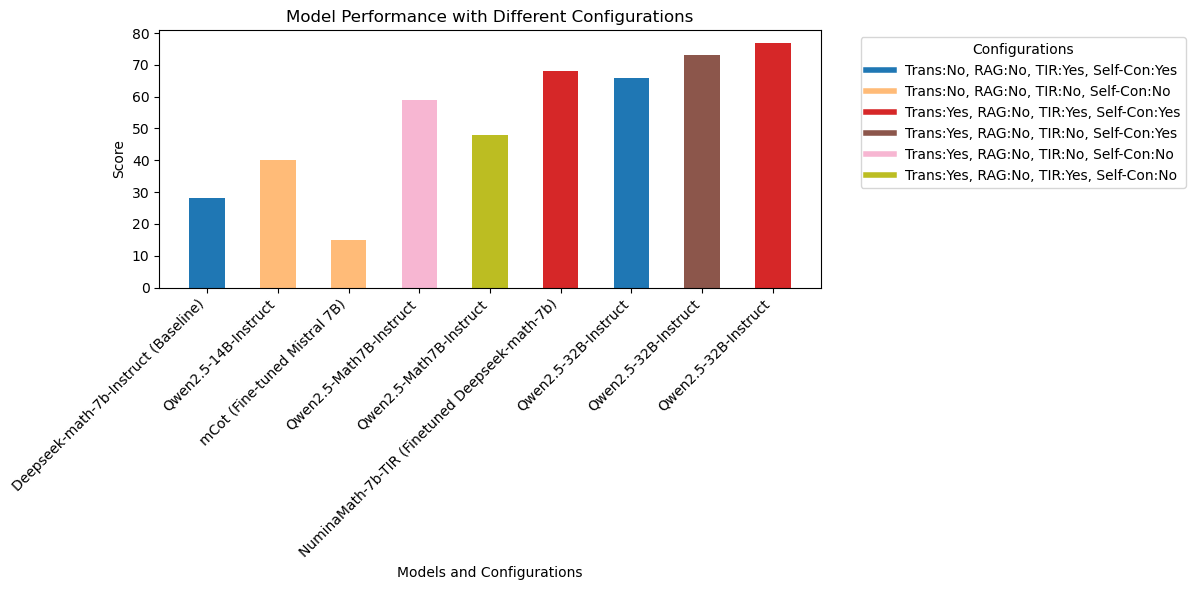}
  \label{fig:graph}
\end{figure*}

\begin{figure*}[ht]
Overall pipeline of our solution
  \centering
  \includegraphics[width=\textwidth]{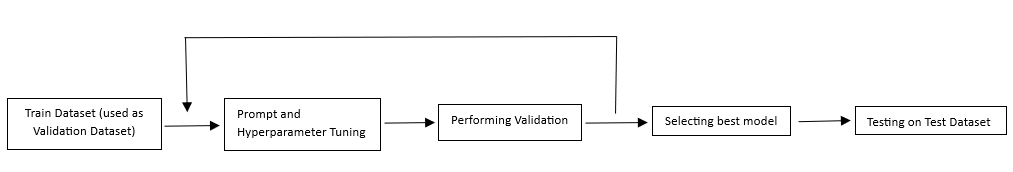}
  \caption{Iterative pipeline}
  \label{fig:pipe}
\end{figure*}
The experiment began with a baseline score of 28, and the fine-tuned Mistral 7B model scored only 15, indicating the need for a more powerful model architecture and translation capabilities. Scaling to larger Qwen 2.5 models, such as the 7B and 14B-Instruct, improved performance to 40 points but still showed limitations with certain problem types. The quantized Qwen 2.5-32B-Instruct model performed better, reaching a score of 49 without translation, benefiting from its larger capacity.

Introducing translation using the 14B-Instruct model boosted the score to 61, allowing the model to leverage pre-existing English knowledge. Further improvement occurred when translation was done using the 32B model. Incorporating Retrieval-Augmented Generation (RAG) with translated English questions raised the score to 70, showing that additional context from external datasets enhanced model performance. However, removing RAG led to a score of 73, suggesting that it may have introduced noise, or the larger model could handle the task independently.

The introduction of Tool Integrated Reasoning (TIR), where Python was used for complex calculations, brought the score to 76. This step significantly enhanced the model's efficiency and accuracy, especially with Bengali questions, which saw a notable improvement. The final model, optimized with prompt engineering and hyperparameter tuning, achieved a score of 77, marking the culmination of the model’s refinement in handling diverse problem types.

\newpage
\section{Discussion}
Throughout the development of our AI model for solving Bengali mathematical problems, we encountered several key challenges and insights:
\subsection{Translation Impact}
Bengali questions presented without translation consistently scored lower than their translated counterparts. Translating questions into English improved model understanding and problem-solving, leveraging the model's richer pre-trained knowledge in English.
\subsection{Model Size for Translation}
Using the Qwen 2.5-32B-Instruct model for translation yielded better results compared to the Qwen 2.5-14B-Instruct model. The larger model demonstrated superior language comprehension, which improved translation quality and contributed to overall score improvements.
\subsection{RAG Limitations}
Initial experiments with Retrieval-Augmented Generation (RAG) aimed to enhance the model’s responses by adding contextual information. However, RAG often introduced noise, leading to a decrease in performance. Removing RAG subsequently improved scores, indicating that the inherent capabilities of the larger Qwen models were sufficient without additional retrieved context for this task.
\subsection{Scarcity of Bengali Datasets}
The limited availability of high-quality Bengali mathematical datasets restricted the model's exposure to diverse problem types. 
\subsection{Inference Optimization with VLLM}
Integrating Variable-Length Language Modeling (VLLM) enabled faster inference, which was crucial for handling the computational demands of larger models. This optimization helped streamline the testing process within our resource constraints
\subsection{Prompt Sensitivity}
The model exhibited a high sensitivity to prompt wording, with minor adjustments in phrasing significantly affecting output quality and accuracy. Effective prompt engineering became essential to guide the model toward optimal solutions, underscoring the importance of precision in prompt construction.

\subsection{Quantization for Resource Constraints}
Due to Kaggle's limited GPU memory, we had to quantize the larger Qwen models to make them feasible for testing. This process reduced memory requirements, allowing us to utilize the larger 32B model in a resource-limited environment without compromising performance excessively.
\subsection{Fine-Tuning Limitations for Larger Models}
While fine-tuning smaller Qwen models was manageable, memory constraints in Kaggle prevented us from fine-tuning the Qwen 2.5-32B model. This limitation hindered further customization of the 32B model, restricting our ability to fine-tune it for specific Bengali problem-solving tasks.

\section{Conclusion and Future Work}
This paper presented a novel approach to solving Bengali mathematical problems by leveraging the Qwen 2.5 series models and optimizing through prompt engineering, translation, and Tool Integrated Reasoning (TIR). Our methods demonstrated the efficacy of advanced NLP techniques in mathematical problem-solving, particularly for low-resource languages like Bengali. The Qwen 2.5-32B model, combined with techniques such as self-consistency and TIR, achieved significant improvements over baseline models, highlighting the model’s potential in handling complex reasoning tasks with minimal Bengali-specific data.  
Despite these achievements, challenges remain, especially regarding translation dependencies, limited Bengali datasets, and memory constraints for large models. These factors impacted both model performance and adaptability to a wider range of problem types. Our results also revealed the sensitivity of model outputs to prompt phrasing, underscoring the need for refined prompt engineering techniques.

To build on this work, we propose the following areas for future exploration:

\begin{itemize}
    \item \textbf{Enhanced Bengali Data Collection:} Increasing the availability and diversity of Bengali mathematical datasets will enable better model training and adaptability to complex problem types.
    \item \textbf{Domain-Specific Fine-Tuning:} With sufficient resources, fine-tuning the Qwen 2.5-32B model specifically on Bengali math problems could further improve accuracy and reasoning ability.
    \item \textbf{Optimized Prompt Engineering:} Researching prompt optimization strategies for mathematical reasoning tasks, especially in low-resource languages, could make the model responses more consistent and accurate.
    \item \textbf{Exploration of Lightweight Models:} Investigating smaller, efficient models tailored for Bengali could balance memory constraints and inference speed without sacrificing accuracy.
\end{itemize}

By addressing these areas, we aim to enhance the utility of deep learning models in Bengali problem-solving, making advanced educational tools more accessible to Bengali-speaking learners and educators. This research lays a foundation for further progress in low-resource language applications of AI in mathematical education and reasoning.


\end{document}